\documentclass{article}

\usepackage{PRIMEarxiv}

\usepackage[utf8]{inputenc}
\usepackage[T1]{fontenc}
\usepackage{hyperref}
\usepackage{url}
\usepackage{booktabs}
\usepackage{amsfonts}
\usepackage{nicefrac}
\usepackage{microtype}
\usepackage{fancyhdr}
\usepackage{graphicx}
\graphicspath{{media/}}

\usepackage[table]{xcolor}
\usepackage{pifont}
\usepackage{colortbl}
\usepackage{array}
\usepackage{subcaption}
\usepackage{multirow}
\usepackage{tabularx}
\usepackage{tcolorbox}
\tcbuselibrary{raster}
\usepackage{placeins}

\newcommand{\cmark}{\ding{51}}
\newcommand{\xmark}{\ding{55}}
\definecolor{narrategreen}{RGB}{198,239,206}
\definecolor{gtcol}{RGB}{214,240,224}
\definecolor{retcol}{RGB}{208,232,248}
\definecolor{ftcol}{RGB}{253,232,228}
\definecolor{zscol}{RGB}{236,224,245}
\newcommand{\best}[1]{\textcolor{green!55!black}{\underline{\textbf{#1}}}}
\newcommand{\worst}[1]{\textcolor{red!70!black}{#1}}
\newcommand{\ms}[2]{#1${}_{{\scriptstyle\pm}#2}$}

\title{NARRATE: A Multimodal Real-World Australian Driving Dataset for Human-Centred Explanations in Automated Driving}

\author{
\textbf{Ashkan Yousefi Zadeh}$^{1,2}$,
\textbf{Zishuo Zhu}$^{1,2}$,
\textbf{Xiaomeng Li}$^{1,2}$,\\
\textbf{Andry Rakotonirainy}$^{1,2}$,
\textbf{Sebastien Glaser}$^{1,2}$,
\textbf{Ronald Schroeter}$^{1,2}$,\\
\textbf{Patricia Delhomme}$^{3}$,
\textbf{Zahra Mehraban}$^{1,2}$\\[4pt]
$^1$Queensland University of Technology (QUT), Faculty of Health,\\
School of Psychology and Counselling, Australia\\
$^2$ARC Training Centre for Automated Vehicles in Rural and Remote Regions (AVR3), Australia\\
$^3$Universit\'e Gustave Eiffel, Laboratory of Applied Psychology and Ergonomics, France
}

\begin{document}
\maketitle

\begin{abstract}
Automated vehicles must explain their decisions in ways that passengers can understand, monitor, and trust. Existing language-annotated driving datasets are mostly observer-written, post-hoc, simulation-based, or generated from sensor inputs, rather than elicited from the driver performing the action. We introduce NARRATE, a multimodal real-world Australian driving dataset comprising 2,050 annotated events from 35 experienced drivers and driving instructors on public roads. Each event is grounded in synchronised visual, localisation, motion, and LiDAR streams and paired with in-vehicle and/or post-drive free-text explanations. NARRATE provides action labels, scenario-context labels spanning six high-level and 32 fine-grained categories, and span-level Situational Awareness (SA) annotations over driver explanations for Perception, Comprehension and Projection. Four benchmark tasks (SA, scenario-context, driver-action classification, and explanation generation) show that this structure is learnable from driver language, while fine-grained context recognition and explanation generation remain challenging. NARRATE paves a path towards more human-centred and domain-aware explanation models for automated driving.
\end{abstract}

\keywords{Multimodal Driving Dataset \and Human-Centred Explanations \and Situational Awareness}

\section{Introduction}

As automated vehicles (AVs) move from controlled test tracks to public roads, explaining driving decisions becomes central to safe and trustworthy human--vehicle interaction~\cite{koo2015why,zhu2025human}. Passengers need to understand not only what the vehicle is doing, but why a manoeuvre is appropriate in the current road situation~\cite{koo2015why,capallera2022human}. Explainable AI (XAI) can make model decisions more interpretable and accountable~\cite{adadi2018peeking}, but explanation quality should not be defined only by researchers' assumptions about what is informative. It should also be grounded in how humans naturally explain events~\cite{miller2019explanation}, particularly in automated driving, where explanation content, timing, modality, and context affect comprehension, workload, and trust~\cite{koo2015why,zhu2025human,yousefi2025psylingxav}. Despite rapid progress in driving datasets paired with natural-language annotations, most do not capture explanations from the person who performed the driving action. Instead, the accompanying language is often written post-hoc by observers, generated or reconstructed from sensor evidence, structured as question--answer pairs, or collected in simulation~\cite{kim2018textual,xu2020explainable,marcu2024lingoqa,sima2024drivelm,nie2024reason2drive,arai2025covla,wang2025omnidrive,ma2024lampilot}. These resources are valuable, but primarily reconstruct plausible reasons from an external viewpoint. They leave open a basic empirical question: how do experienced drivers explain their own decisions in real traffic, and what situational information do they include? In this paper, \emph{human-centred} explanations originate from the driver performing the task, rather than from external observers or post-hoc annotators.

\begin{figure}[t]
\centering
\includegraphics[width=\linewidth]{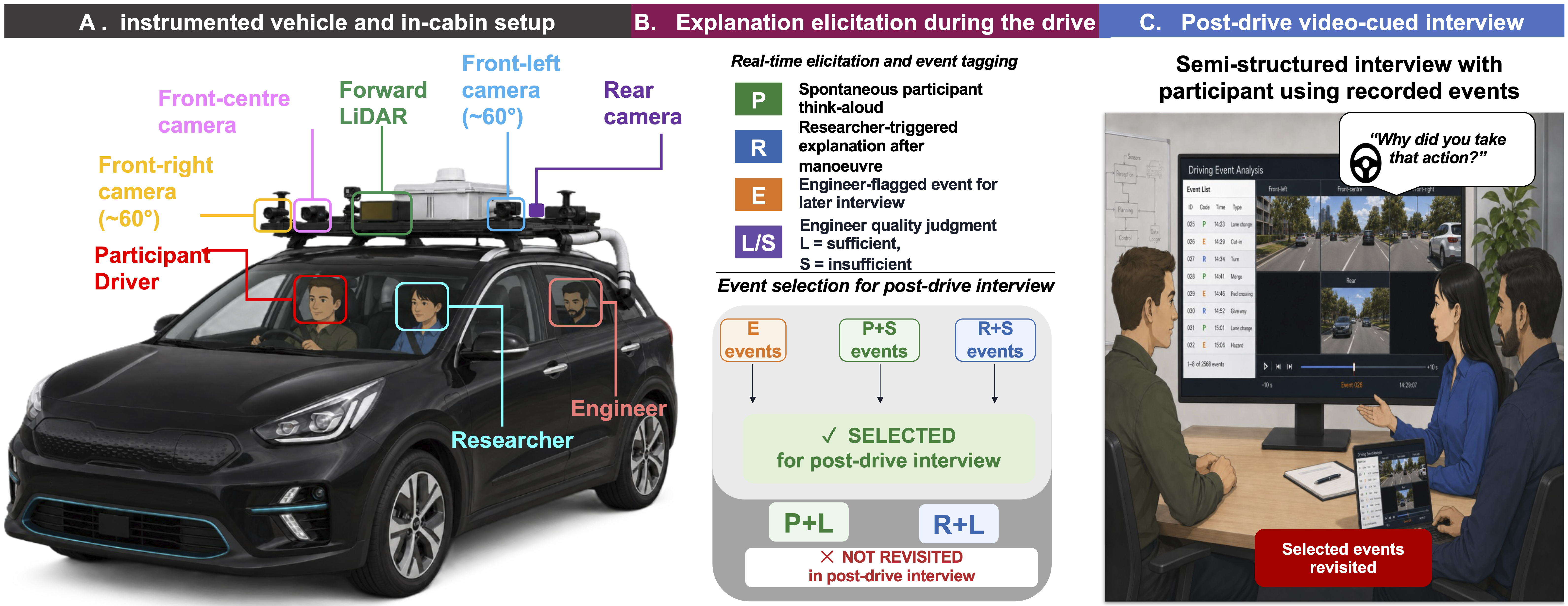}
\caption{NARRATE data collection protocol:
(A)~instrumented vehicle and in-cabin setup;
(B)~real-time explanation elicitation and event tagging during the drive;
(C)~post-drive video-cued interview using the synchronised multi-view
replay dashboard.}
\label{fig:data_collection_protocol}
\end{figure}

Driver explanations can go beyond describing an action: drivers may refer to relevant cues, interpret their meaning for the current manoeuvre, and anticipate how the situation may unfold. This corresponds closely to Endsley's Situational Awareness (SA) framework, which distinguishes Perception, Comprehension, and Projection in human performance in dynamic systems~\cite{endsley1995toward,endsley2000situation}. We do not assume that every effective explanation must contain all three SA levels; rather, SA provides a cognitively grounded lens for analysing which aspects of situational reasoning are expressed in driver-produced explanations, complementing work showing that explanations can support understanding, trust calibration, and situation awareness in automated driving~\cite{koo2015why,capallera2022human,zhu2025human}. 

Explanation timing also matters: in-vehicle explanations capture what is salient under real-time workload, whereas post-drive video-cued reflections allow drivers to elaborate after the event. A dataset for human-centred driving explanations should capture both, rather than treating driver reasoning as a single static text label.

\begin{figure}[t]
\centering
\includegraphics[width=\textwidth]{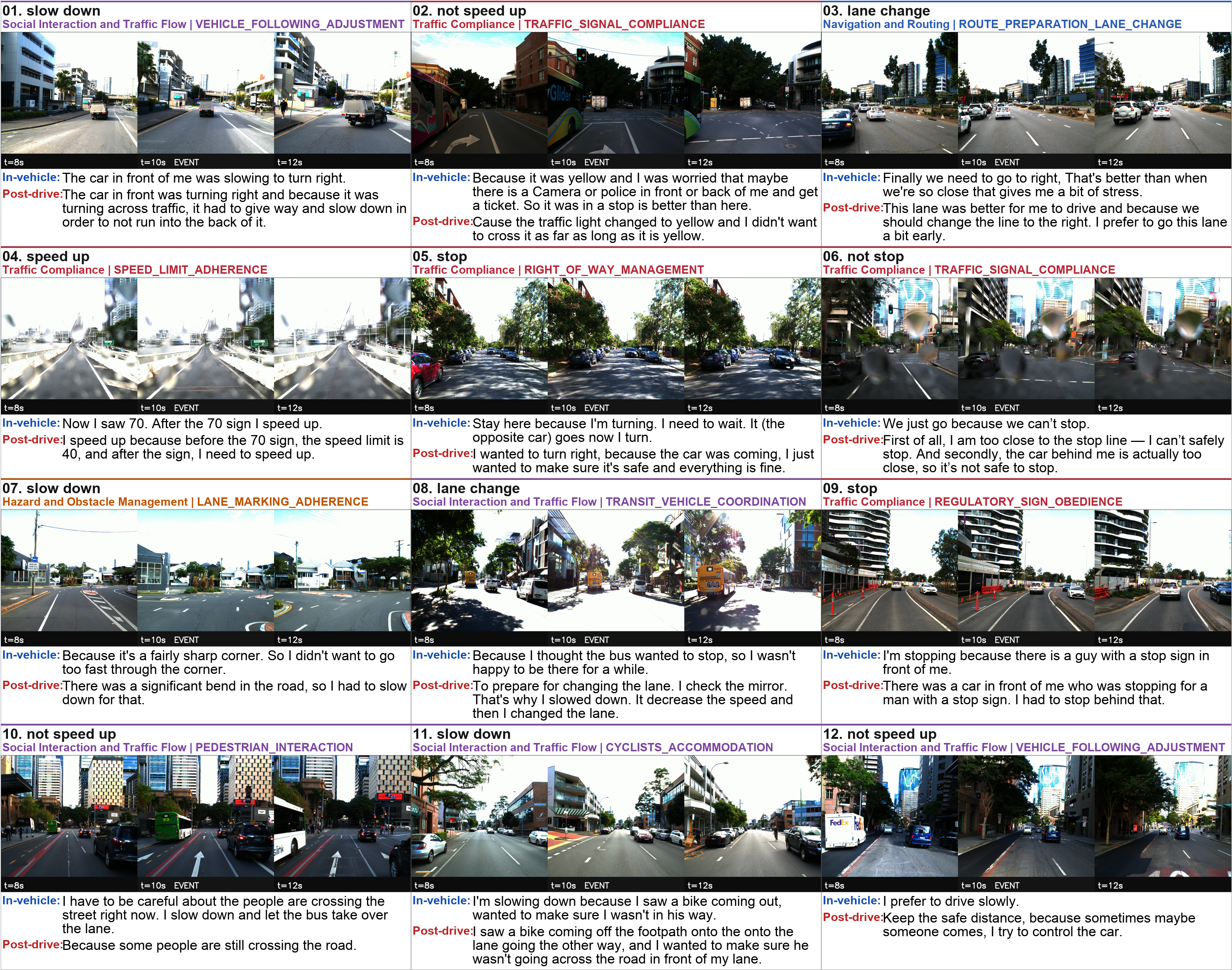}
\caption{
Representative events from NARRATE. Each panel shows an annotated driving event with its driver-action label, high-level and fine-grained scenario-context tags, three front-centre camera frames sampled around the tagged event time, and the corresponding driver explanations. The middle frame marks the tagged event, while the neighbouring frames provide pre- and post-event visual context. In-vehicle explanations were recorded during driving, whereas post-drive explanations were recalled during the video-cued debrief. 
}
\label{fig:examples}
\end{figure}

A further gap is geographic and operational coverage. Most large driving datasets are collected in North American, European, Chinese, or simulated settings, whereas Australian driving involves left-hand traffic, right-hand-drive vehicles, domain-specific road rules, signage, lane-use conventions, and road-user expectations. Recent left-hand-driving domain adaptation work shows that transferring autonomous-steering models from U.S. right-hand-driving data to real-world Australian data requires explicit treatment of domain shift~\cite{mehraban2025saliency}. Explanation datasets should therefore be domain-aware: an AV should explain its behaviour using road conventions and situational cues from its operating environment.

To address these gaps, we introduce \textbf{NARRATE}: Naturalistic Action Reasoning and Real-World Awareness Through Explanations. NARRATE is a multimodal real-world Australian driving dataset designed to help train and evaluate automated driving explanation models that produce human-centred explanations by learning from experienced drivers' own decision explanations. It comprises 2,050 annotated events from 35 experienced drivers and driving instructors on public roads in Brisbane, Queensland, paired with in-vehicle and/or post-drive free-text explanations collected through the dual-timing protocol in Fig.~\ref{fig:data_collection_protocol}. NARRATE provides driver-action labels, scenario-context labels, span-level SA annotations, and benchmark tasks; representative annotated events are shown in Fig.~\ref{fig:examples}.

Our contributions are threefold: (i) a human-centred dataset of real-world driving explanations produced by experienced drivers during and after driving. (ii) Multimodal event grounding with driver-action labels, scenario-context labels across six high-level and 32 fine-grained categories, and span-level SA labels for L1 Perception, L2 Comprehension, and L3 Projection. (iii) Participant-disjoint benchmarks for SA classification, context classification, driver-action classification, and explanation generation. NARRATE is not a fleet-scale perception dataset, but a high-fidelity explanation dataset for studying driver-produced explanations of real-world driving decisions.

\section{Related Work}
\label{sec:related}

Large-scale driving datasets have driven substantial progress in AV perception, localisation, prediction, and planning. nuScenes~\cite{caesar2020nuscenes} established a widely used multi-sensor benchmark combining multi-camera video, LiDAR, radar, GPS, and IMU, while BDD100K~\cite{yu2020bdd100k} demonstrated the value of large-scale front-view video for diverse visual driving tasks. Behaviour-oriented datasets such as HDD~\cite{ramanishka2018toward} go beyond perception by annotating driver actions, goals, causes, and attention-related labels in naturalistic driving. However, these datasets provide perceptual or behavioural supervision rather than natural-language explanations from the driver who performed the manoeuvre.

\begin{table}[tb]
\caption{Comparison with related datasets.
  \textbf{Columns:} Cam.\,=\,number of cameras; Lang.\,=\,language format;
  SA\,=\,Situational Awareness; Act.\,=\,action labels; Ctx.\,=\,context labels;
  Size\,=\,headline annotated count (unit as reported); Hrs\,=\,hours of driving/video;
  Drv\,=\,distinct human drivers.
  \textbf{Sensors:} V\,=\,video; L\,=\,LiDAR; I\,=\,IMU; G\,=\,GPS; C\,=\,CAN;
  Ra\,=\,radar; D\,=\,dynamics; Sim.\,=\,simulator; var.\,=\,variable; mul.\,=\,multiple.
  \textbf{Language:} FT\,=\,free text; QA\,=\,question--answer; Cmd.\,=\,commands;
  Cat.\,=\,categories; Lbl\,=\,labels; CF\,=\,counterfactual; CoC\,=\,chain of causation.
  \textbf{Scale:} n/r\,=\,not reported; n/a\,=\,not applicable (built on another dataset
  or simulated); crowd\,=\,crowdsourced (no fixed driver count); der.\,=\,derived by us;
  seg.\,=\,segments; scn.\,=\,scenes; sess.\,=\,sessions.
  $\ddagger$\,Reasoning structure, not Endsley L1/L2/L3.
  $*$\,Causal/scenario labels only.
  $\dagger$\,Instruction or advice, not driver explanation.
  \textsuperscript{a}\,Alpamayo hours n/r for the 700K CoC set; the paper's 80{,}000\,h
  is the fleet pre-training pool, not annotated reasoning hours.
  \textsuperscript{b}\,CoVLA also stated as 6M frames / 431K sampled trajectory points;
  83.3\,h $=$ 30\,s\,$\times$\,10{,}000.
  \textsuperscript{c}\,HAD: 45{,}629 advice annotations (25{,}549 action / 20{,}080
  attention) over the clips.
  \textsuperscript{d}\,NARRATE Hrs given as annotated\,/\,recorded:
  8.5\,h annotated event footage (2{,}050\,$\times$\,15\,s) and
  $\sim$23\,h total driving recorded (35\,$\times$\,40\,min).}
\label{tab:dataset_comparison}
\centering
\footnotesize
\setlength{\tabcolsep}{10pt}
\renewcommand{\arraystretch}{0.92}
\resizebox{\textwidth}{!}{%
\begin{tabular}{lclllccc rll}
\toprule
\textbf{Dataset} & \textbf{Cam.} & \textbf{Sensors} &
\textbf{Source} & \textbf{Lang.} & \textbf{SA} & \textbf{Act.} & \textbf{Ctx.} &
\textbf{Size} & \textbf{Hrs} & \textbf{Drv} \\
\midrule
\multicolumn{11}{l}{\textit{Perception \& behaviour}} \\
nuScenes~\cite{caesar2020nuscenes}      & 6    & V,L,Ra,G,I  & None              & --               & \xmark              & \xmark & \xmark & 1{,}000 scn.        & 5.5           & n/r \\
BDD100K~\cite{yu2020bdd100k}            & 1    & V,G,I       & None              & --               & \xmark              & \xmark & \xmark & 100K videos         & $\sim$1{,}100 der. & crowd \\
HDD~\cite{ramanishka2018toward}         & 3    & V,L,I,G,C   & Labels only       & Lbl              & \xmark              & \cmark & \cmark$^{*}$ & 137 sess.      & 104           & n/r \\
\midrule
\multicolumn{11}{l}{\textit{Language \& explanation}} \\
Talk2Car~\cite{deruyttere2019talk2car}  & 6    & V,L,Ra,G    & Human commands    & Cmd.$^{\dagger}$ & \xmark              & \cmark & \xmark & 11{,}959 cmds.      & n/a           & n/a \\
HAD~\cite{kim2019grounding}             & 3    & V,C         & Human advice      & FT$^{\dagger}$   & \xmark              & \cmark & \cmark$^{*}$ & 5{,}675 clips\textsuperscript{c} & $\sim$32   & n/a \\
BDD-X~\cite{kim2018textual}             & 1    & V,G,D       & Crowd, post-hoc   & FT               & \xmark              & \cmark & \xmark & 26{,}228 annot.     & $\sim$77      & n/a \\
BDD-OIA~\cite{xu2020explainable}        & 1    & V           & Crowd, post-hoc   & Cat.             & \xmark              & \cmark & \cmark$^{*}$ & 22{,}924 clips & $\sim$32 der. & n/a \\
LingoQA~\cite{marcu2024lingoqa}         & 1    & V           & Human + GPT       & QA               & \xmark              & \cmark & \xmark & 419.9K QA           & n/r           & n/r \\
DriveLM~\cite{sima2024drivelm}          & 6    & V,L,Ra      & Semi-auto + human & Graph QA         & \xmark$^{\ddagger}$ & \cmark & \xmark & $\sim$1.6M QA       & n/r           & n/a \\
Reason2Drive~\cite{nie2024reason2drive} & var. & V,L         & Auto + GPT-4      & Chain QA         & \xmark$^{\ddagger}$ & \cmark & \xmark & 632{,}955 QA        & n/r           & n/a \\
CoVLA~\cite{arai2025covla}              & 1    & V,I,G,C,D   & Auto-generated    & FT               & \xmark              & \cmark & \cmark$^{*}$ & 10{,}000 scn.\textsuperscript{b} & 83.3      & n/r \\
OmniDrive~\cite{wang2025omnidrive}      & 6    & V,L,Ra      & Rule+GPT+human    & QA+CF            & \xmark$^{\ddagger}$ & \cmark & \xmark & nuScenes keyfr.     & n/r           & n/a \\
LaMPilot~\cite{ma2024lampilot}          & --   & Sim.        & Semi-human        & Cmd.$^{\dagger}$ & \xmark              & \cmark & \cmark$^{*}$ & 4{,}900 scn.   & n/r           & n/a \\
Alpamayo-R1~\cite{wang2025alpamayo}     & mul. & V,D         & Auto + human      & CoC$^{\ddagger}$ & \xmark$^{\ddagger}$ & \cmark & \cmark$^{*}$ & 700K seg.\textsuperscript{a} & n/r\textsuperscript{a}  & n/r \\
\midrule
\rowcolor{narrategreen}
\textbf{NARRATE (ours)}                 & \textbf{4} & \textbf{V$\times$4,L,I,G,D}
& \textbf{Experienced drivers, dual-timing}
& \textbf{FT (dual)} & \cmark & \cmark & \cmark
& \textbf{2{,}050 events} & \textbf{8.5\,/\,23}\textsuperscript{d} & \textbf{35} \\
\bottomrule
\end{tabular}%
}
\end{table}

A second line of work introduces language into driving, but with different objectives and language sources. Talk2Car~\cite{deruyttere2019talk2car} grounds passenger commands to objects in driving scenes, HAD~\cite{kim2019grounding} provides human advice for driving videos, and LaMPilot~\cite{ma2024lampilot} studies language-program instruction following in simulation. Explanation datasets such as BDD-X~\cite{kim2018textual} and BDD-OIA~\cite{xu2020explainable} link actions to explanatory text or categories, but use post-hoc observer explanations rather than driver-produced ones. Recent reasoning datasets, including LingoQA~\cite{marcu2024lingoqa}, DriveLM~\cite{sima2024drivelm}, Reason2Drive~\cite{nie2024reason2drive}, OmniDrive~\cite{wang2025omnidrive}, CoVLA~\cite{arai2025covla}, and Alpamayo-R1~\cite{wang2025alpamayo}, advance language-based driving reasoning through QA, counterfactual, causal, or generated supervision. Yet their language is typically generated, reconstructed, or annotated after the fact from sensor evidence, not elicited from the person who made the driving decision. Table~\ref{tab:dataset_comparison} summarises these differences in sensor modalities, language source, annotation format, and action, context, and SA supervision. NARRATE combines real-world multi-sensor grounding, experienced-driver free-text explanations, action and scenario-context labels, and explicit span-level SA annotation.

Building on the SA motivation above, NARRATE also differs in how it represents situational reasoning. Originally developed for human performance in dynamic environments~\cite{endsley1995toward}, SA is relevant to automated driving because reduced driver engagement can weaken situation awareness and takeover readiness~\cite{endsley2017here,merat2009drivers}. It is central to human--vehicle interfaces that communicate what the vehicle perceives, how it interprets the scene, and what risks may follow~\cite{capallera2022human,white2019rebuilding}, and is linked to trust calibration~\cite{petersen2019situational}. Existing driving reasoning datasets may include perception-, prediction-, planning-, or causal-reasoning structures, but not span-level Endsley L1/L2/L3 annotation grounded in driver-produced language. NARRATE complements them with span-level SA labels over experienced-driver explanations, enabling models to learn both what drivers do and how their explanations express perception, comprehension, and projection.

\section{Data Collection}
\label{sec:collection}

\noindent\textbf{Instrumented vehicle.}
Data were collected using a Kia Niro SUV with a roof-mounted sensor rig and onboard edge-compute unit running ROS~2 (Humble)~\cite{macenski2022ros2} (Fig.~\ref{fig:data_collection_protocol}). All streams were recorded as ROS~2 bag files and time-synchronised via GPS pulse-per-second (PPS) through the inertial navigation system. For each event, the platform captured four lossless camera views at 10 fps: front-centre, forward-left, forward-right, and rear-centre, yielding 8,200 camera sequences across 2,050 events. The vehicle also recorded a Leishen CH128X forward-facing LiDAR ($120^{\circ}{\times}25^{\circ}$ field of view, 160m range, 128 channels), GPS-aided INS/IMU streams with PPS synchronisation, GNSS/GPS position, vehicle speed, and longitudinal acceleration.

\noindent\textbf{Participants.}
Thirty-seven participants were recruited. Two sessions were excluded before dataset construction. One lacked recoverable event timestamps needed to extract time-aligned sensor clips and explanations, and the other involved non-protocol-compliant driving. The final analysed sample comprised 35 participants: 30 experienced non-instructor drivers and 5 licensed driving instructors. All included participants were at least 30 years old, held a Queensland open driver's licence for at least three years, were familiar with Brisbane CBD, and reported driving at least two hours per week. Driving instructors additionally held a valid instructor licence for at least one year, and individuals involved in an at-fault crash within the previous two years were excluded. The final sample comprised 24 male and 11 female participants, with a mean age of 44.5 years (SD=12.6; range 30--81) and a mean of 23.3 years of driving experience (SD=14.4; range 3--64). All participants provided written informed consent.

\noindent\textbf{Driving protocol.}
Each session comprised approximately 40 minutes of naturalistic driving on public roads in Brisbane. Participants followed a predefined route on an in-vehicle navigation display; before driving, they were briefed on the route, instructed to drive normally while obeying road rules, and reminded to prioritise safe vehicle control. This standardised exposure to the same road infrastructure while preserving natural variation in traffic, road-user behaviour, and signal timing. Route composition and road-feature statistics are reported in Sec.~\ref{sec:statistics}. Data were collected between approximately 7 am and 5 pm, under mostly sunny conditions with some cloudy and rainy sessions. Driving was manual, without Advanced Driver Assistance Systems (ADAS) intervention, in a left-hand-traffic, right-hand-drive Australian road context. Three people occupied the vehicle: the participant driver, a front-seat researcher, and a rear-seat data engineer. The engineer monitored sensor recording and tagged candidate events in real time; the researcher elicited in-vehicle explanations only after completed actions and only when safe.

\noindent\textbf{Explanation elicitation.}
NARRATE was designed to capture both immediate and reflective driver explanations. During the drive, events arose from participant think-aloud comments (\texttt{P}), researcher-triggered prompts after completed actions (\texttt{R}), or engineer-flagged deferrals for events unsafe or impractical to discuss while driving (\texttt{E}). For \texttt{P} and \texttt{R} events, the engineer marked whether the in-vehicle explanation was sufficiently informative: short or insufficient explanations were marked \texttt{S}, and informative ones \texttt{L}. Events coded \texttt{P+S} or \texttt{R+S}, together with all \texttt{E}-tagged events, were shown to the participant during the post-drive interview to elicit or clarify the explanation; \texttt{P+L} and \texttt{R+L} events were not shown again. After the drive, participants completed a video-cued semi-structured interview using a purpose-built multi-view replay dashboard. Each event was represented by a 15-second synchronised four-view clip, with the tagged manoeuvre at approximately 10 seconds, providing roughly 10 seconds of pre-event context and 5 seconds of post-event outcome. For each post-drive event, participants were asked up to four pre-specified questions: what action they took, why they took it, what they perceived, and what could have happened otherwise. When a driver did not perform an expected manoeuvre, questions used a why-not framing. These omitted-manoeuvre cases are preserved as distinct non-action classes: \emph{not lane change}, \emph{not stop}, \emph{not speed up}, and \emph{not slow down}, not conflated with counterfactual responses. Explanation audio was transcribed using automatic speech recognition, manually corrected, and checked during annotation.

\section{Annotation Design}
\label{sec:annotation}

NARRATE provides two complementary annotation layers: span-level Situational Awareness (SA) labels over driver explanations and event-level scenario-context labels, describing both the reasoning expressed in language and the driving situation.

\noindent\textbf{Situational Awareness annotation.}
SA labels follow Endsley's three-level framework~\cite{endsley1995toward}: \textit{L1 Perception} denotes information the driver noticed, \textit{L2 Comprehension} its driving relevance, and \textit{L3 Projection} anticipation of a future state or risk. Annotators labelled all SA levels expressed in each explanation and highlighted supporting spans. In-vehicle and post-drive explanations were annotated independently, and a single explanation could contain multiple SA levels. For example, in ``The light ahead was turning amber [L1], so I knew I needed to slow down [L2], because there was a car close behind me that might not stop in time [L3]'', all three levels appear in one explanation. This span-level design identifies whether an SA level is present and where perceptual, interpretive, and anticipatory reasoning appears in driver-produced language.

\noindent\textbf{Scenario-context annotation.}
Context labels were assigned at the event level using a driving-scenario taxonomy from prior work~\cite{zadeh2026x}. Each event could receive multiple fine-grained labels from 32 scenario categories grouped into six high-level contexts: Traffic Compliance, Social Interaction and Traffic Flow, Navigation and Routing, Hazard and Obstacle Management, Special Zones and Stops, and Environmental and Adaptation. Context was annotated at the event level rather than the span level because it may depend on the explanation, visual scene, manoeuvre, and surrounding traffic situation.

\noindent\textbf{Annotators and label resolution.}
Three experienced annotators (A, B, C), with complementary expertise spanning automated driving research, natural language processing, computer vision, and human factors, contributed to the annotation process. Annotator~A labelled the full dataset for both SA and context. To estimate reliability, a 303-instance subset (covering approximately 15\% of annotated events and all 35 participants) was selected by stratified sampling to match the full-dataset distribution across high-level context categories and train/validation/test splits~\cite{krippendorff2004content} (Fig.~\ref{fig:sample_dist}), ensuring that agreement estimates were representative of the full corpus rather than an artefact of sample composition. Annotators~B and~C independently labelled SA on this subset, and Annotator~C additionally labelled context. All annotators followed the same guideline document, worked independently in Label Studio~\cite{labelstudio}, and were blinded to each other's labels. They could view the synchronised event video alongside the explanation text, so labels were grounded in both language and driving context. For the reliability subset, L1 and L2 SA labels were resolved by majority vote across Annotators~A, B, and~C. L3 SA labels and context labels were resolved by consensus between Annotators~A and~C, as Annotator~B's markedly lower L3 positive rate (13.2\%) relative to A (71.9\%) and C (59.7\%) indicated a systematically different interpretation of projection that would bias majority-vote outcomes at that level. For the remaining 85\% of events, the released labels follow Annotator~A's guideline-based annotations; the substantial A--C agreement at L3 ($\kappa{=}0.690$) reported in Sec.~\ref{sec:statistics} supports the reliability of those labels. Agreement results are reported in Sec.~\ref{sec:statistics}.

\section{Dataset Statistics}
\label{sec:statistics}

\noindent\textbf{Scale, splits, and sensor coverage.}
From 2,138 raw event tags, 85 events were excluded because the event tag was erroneous (e.g., an accidental or incorrectly timed tag), the response was irrelevant, the participant could not recall the event, or the event could not be reliably located in the transcript. A further 3 events had no valid explanation, yielding 2,050 annotated events from 35 participants. Events were partitioned into participant-disjoint train, validation, and test splits using stratified group splitting to balance action, context, and SA label distributions. Table~\ref{tab:dataset} reports the resulting split sizes.

\begin{table}[tb]
\centering
\footnotesize
\setlength{\tabcolsep}{10pt}
\renewcommand{\arraystretch}{0.92}
\caption{NARRATE dataset splits.
IV = in-vehicle; Post = post-drive.}
\label{tab:dataset}
\begin{tabular}{@{}lrrrr@{}}
\toprule
\textbf{Component} & \textbf{Train} & \textbf{Val} & \textbf{Test} & \textbf{Total} \tabularnewline
\midrule
Participants             & 24      & 5   & 6   & 35      \tabularnewline
Events                   & 1{,}402 & 272 & 376 & 2{,}050 \tabularnewline
IV explanations          & 875     & 171 & 227 & 1{,}273 \tabularnewline
Post explanations        & 750     & 155 & 205 & 1{,}110 \tabularnewline
Dual (both)              & 223     & 54  & 56  & 333     \tabularnewline
IV words (mean$\pm$SD)   & $17.0{\pm}11.4$ & $11.8{\pm}9.5$  & $12.9{\pm}7.9$  & $15.6{\pm}10.8$ \tabularnewline
Post words (mean$\pm$SD) & $26.5{\pm}24.3$ & $21.2{\pm}15.2$ & $18.7{\pm}13.4$ & $24.3{\pm}21.8$ \tabularnewline
\bottomrule
\end{tabular}
\end{table}

\begin{figure}[!t]
\centering
\captionsetup[subfigure]{font=scriptsize,skip=1pt}

\begin{subfigure}[t]{0.305\textwidth}
\centering
\includegraphics[width=\linewidth]{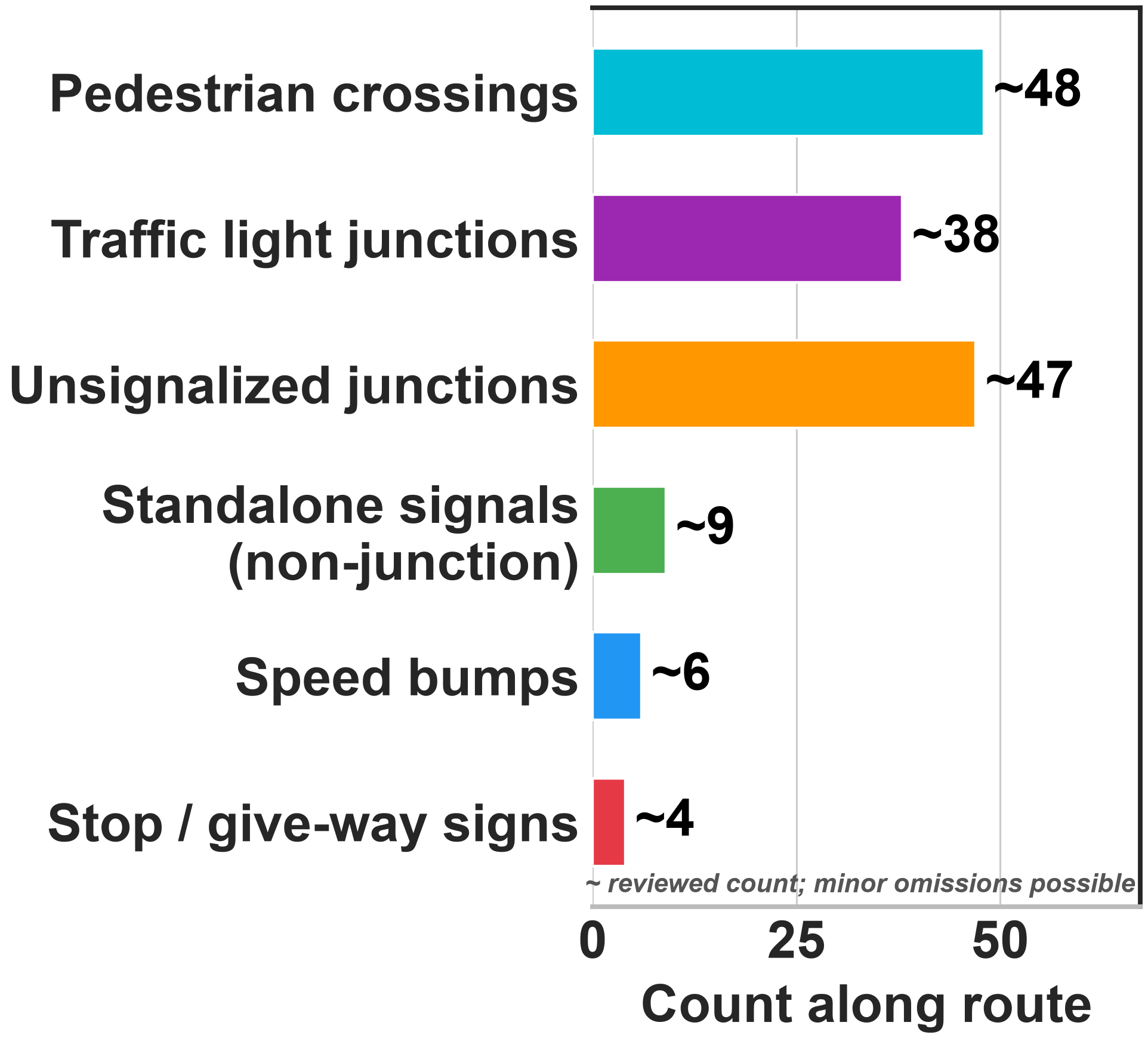}
\caption{Traffic/safety features}
\label{fig:road_features}
\end{subfigure}
\hfill
\begin{subfigure}[t]{0.305\textwidth}
\centering
\includegraphics[width=\linewidth]{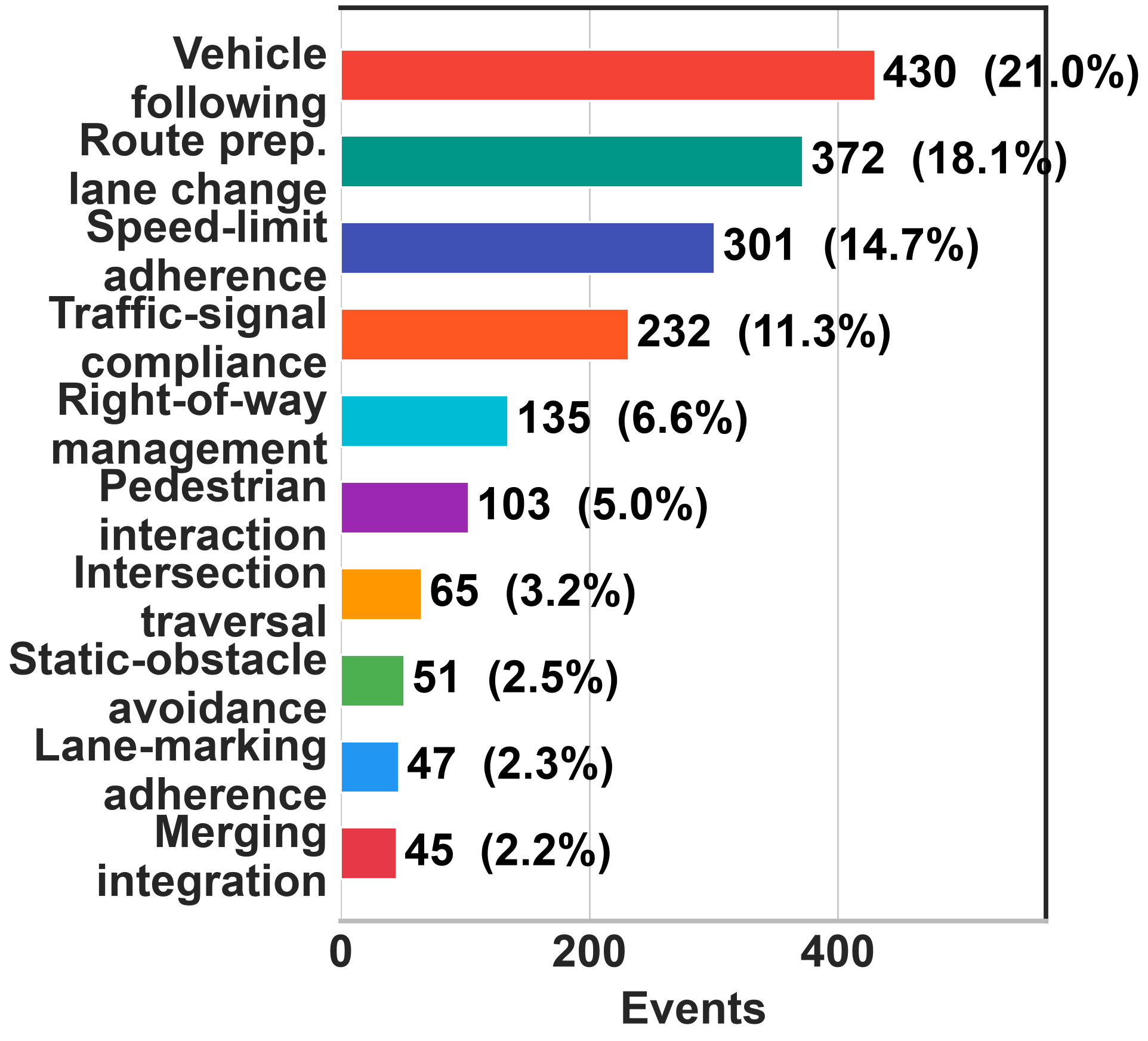}
\caption{Top 10 fine-grained context}
\label{fig:fg_context}
\end{subfigure}
\hfill
\begin{subfigure}[t]{0.305\textwidth}
\centering
\includegraphics[width=\linewidth]{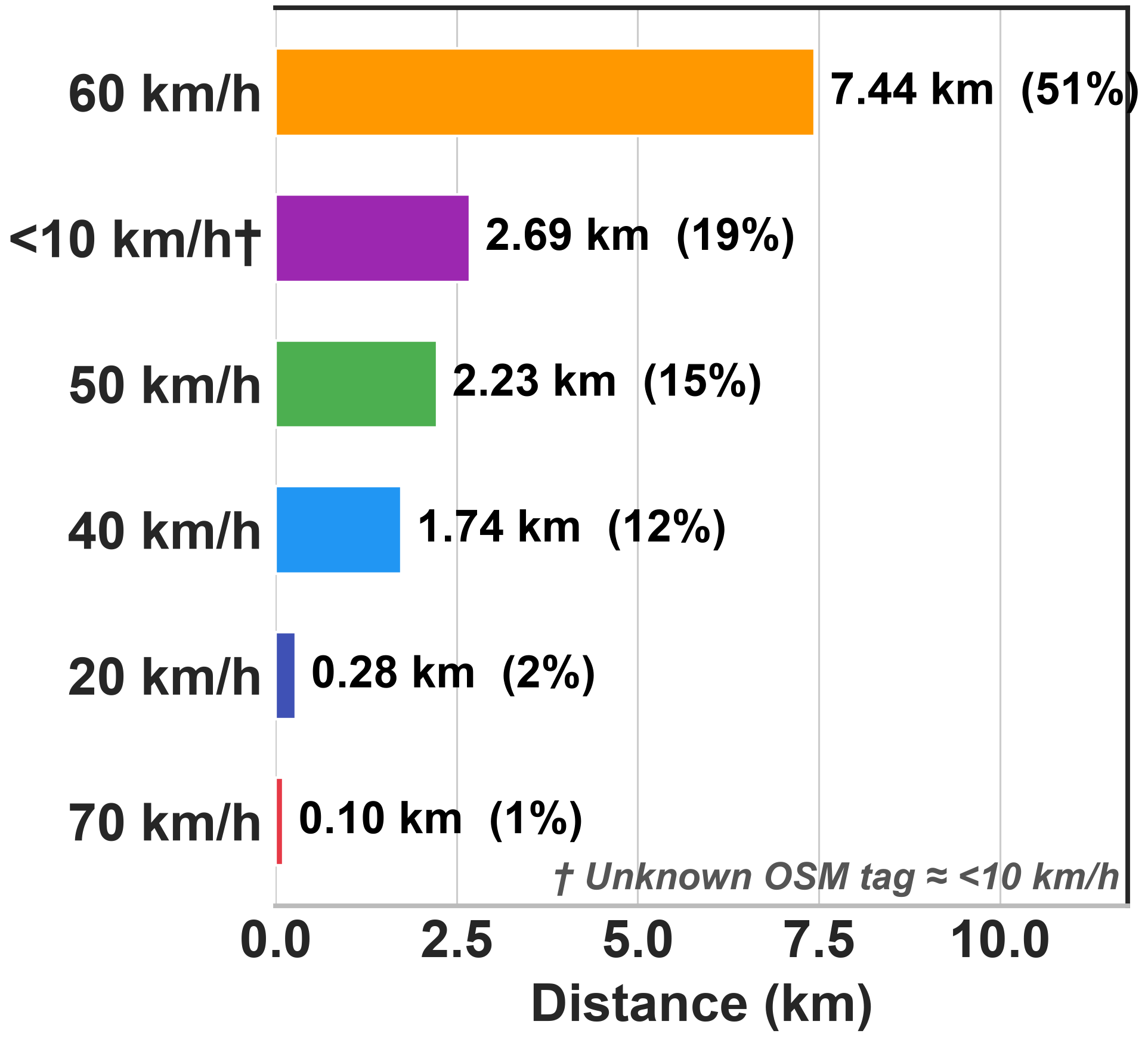}
\caption{Speed zones}
\label{fig:road_speed}
\end{subfigure}

\vspace{2pt}

\begin{subfigure}[t]{0.305\textwidth}
\centering
\includegraphics[width=\linewidth]{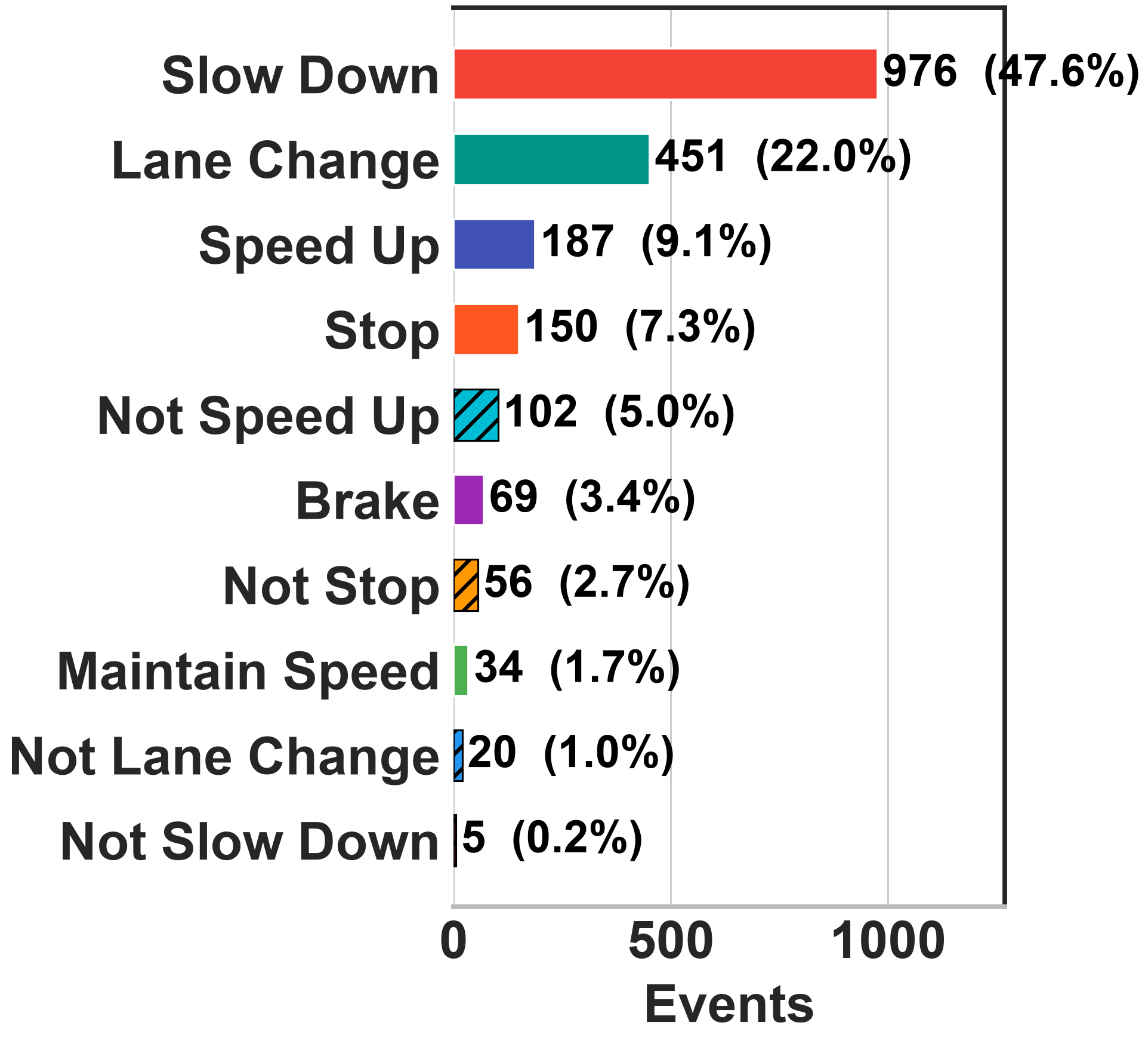}
\caption{Driver action}
\label{fig:action_dist}
\end{subfigure}
\hfill
\begin{subfigure}[t]{0.305\textwidth}
\centering
\includegraphics[width=\linewidth]{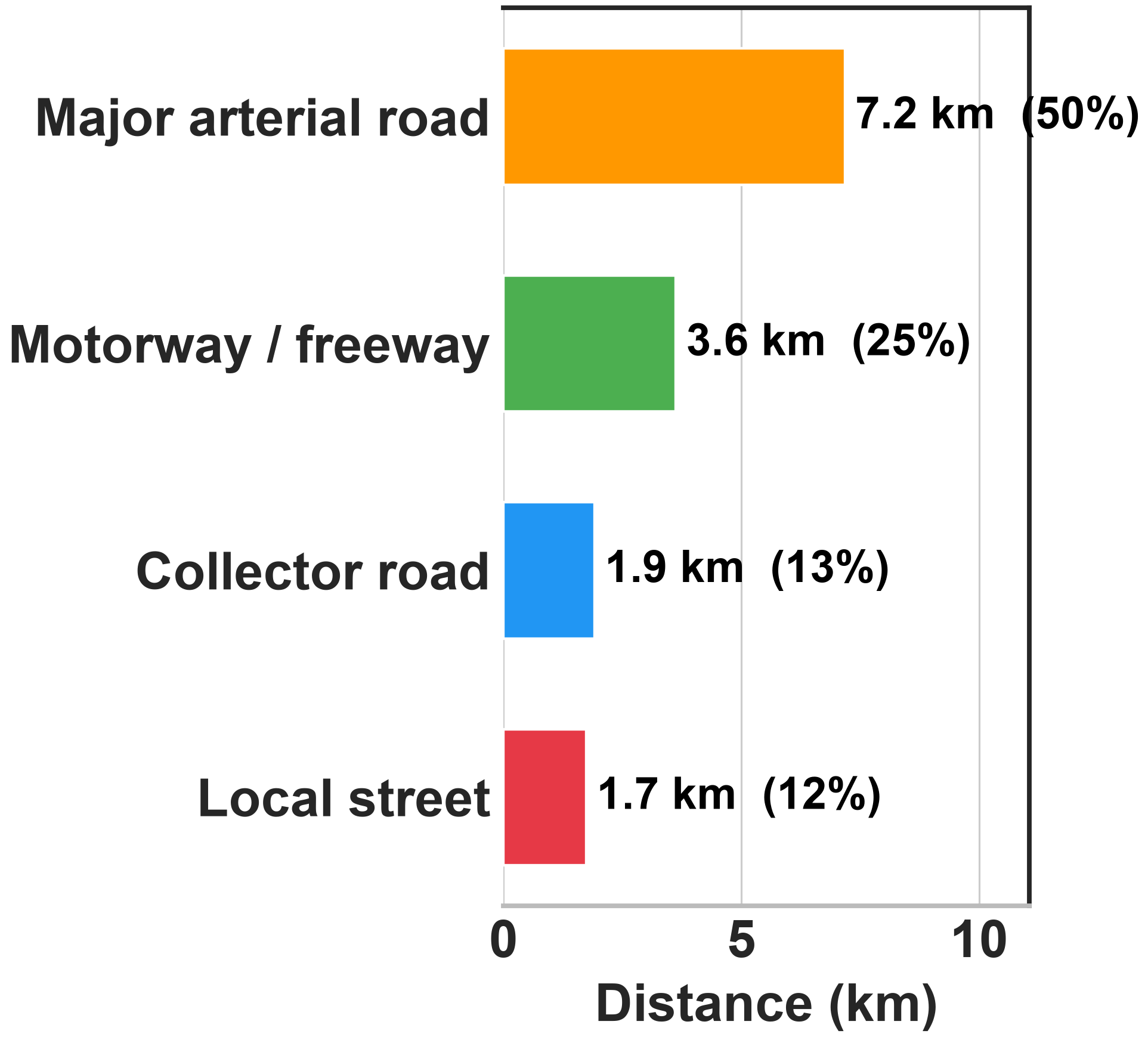}
\caption{Road type}
\label{fig:road_type}
\end{subfigure}
\hfill
\begin{subfigure}[t]{0.305\textwidth}
\centering
\includegraphics[width=\linewidth]{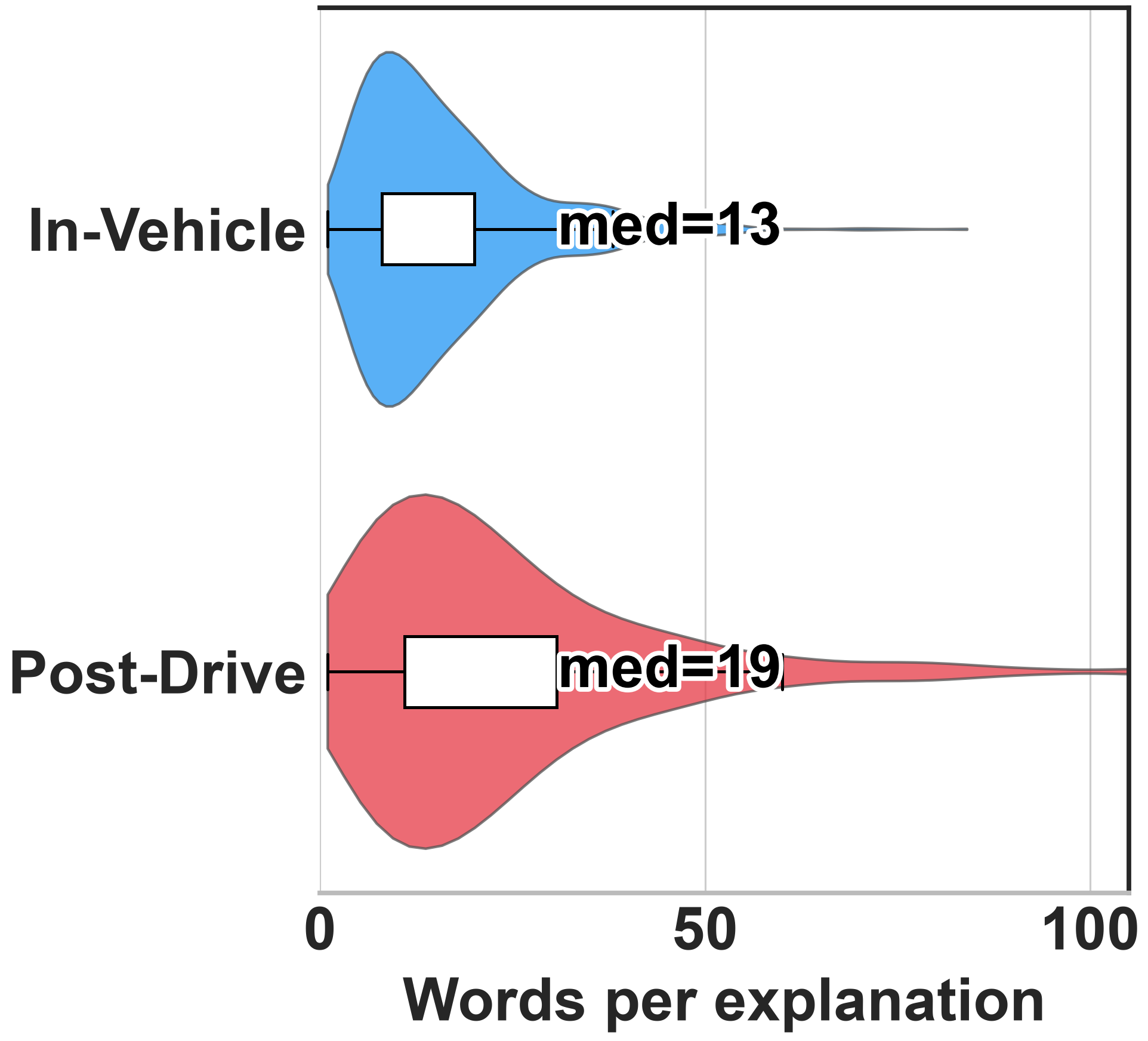}
\caption{Explanation length}
\label{fig:exp_length}
\end{subfigure}

\vspace{2pt}

\begin{subfigure}[t]{0.305\textwidth}
\centering
\includegraphics[width=\linewidth]{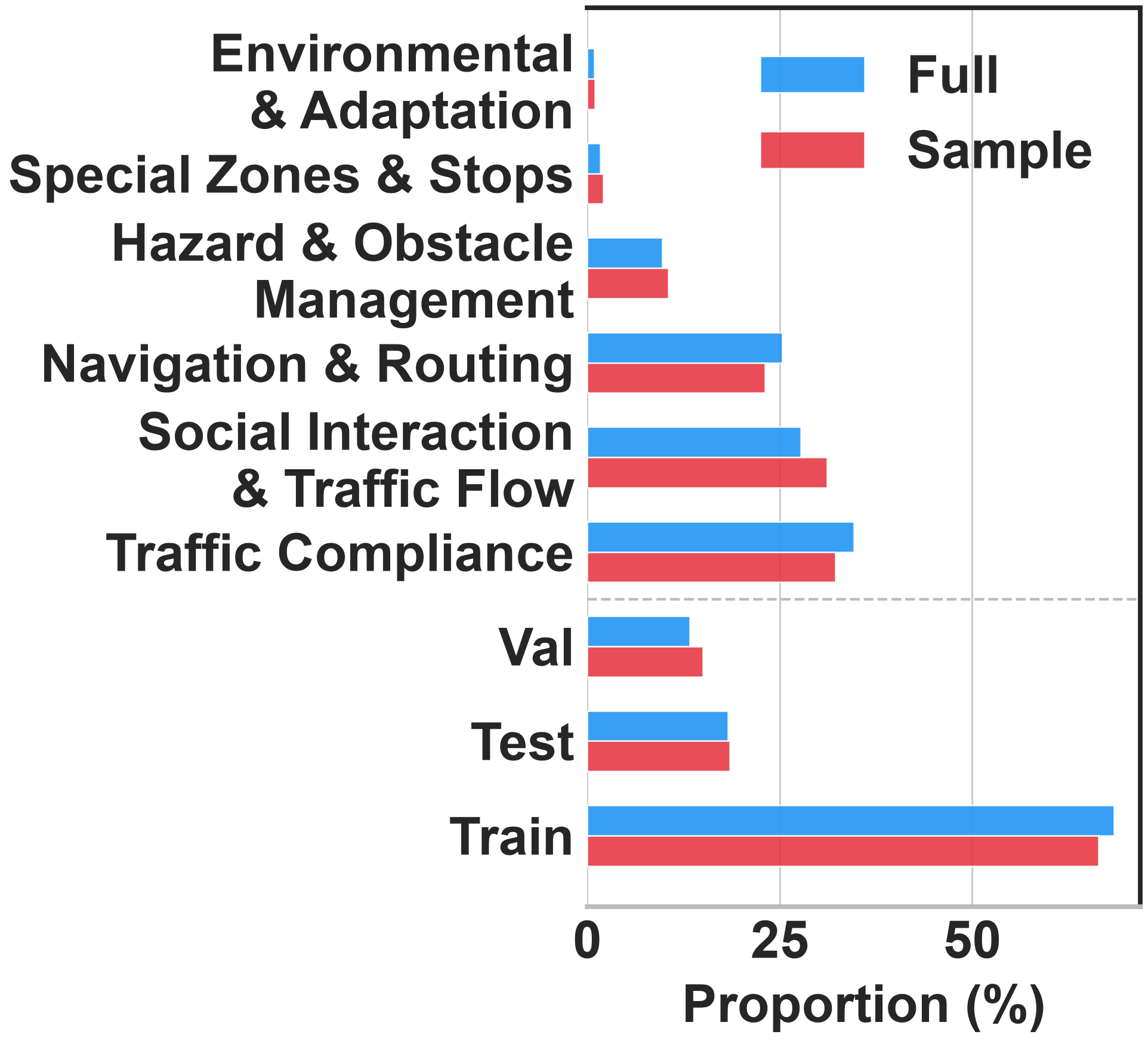}
\caption{Reliability sample}
\label{fig:sample_dist}
\end{subfigure}
\hfill
\begin{subfigure}[t]{0.305\textwidth}
\centering
\includegraphics[width=\linewidth]{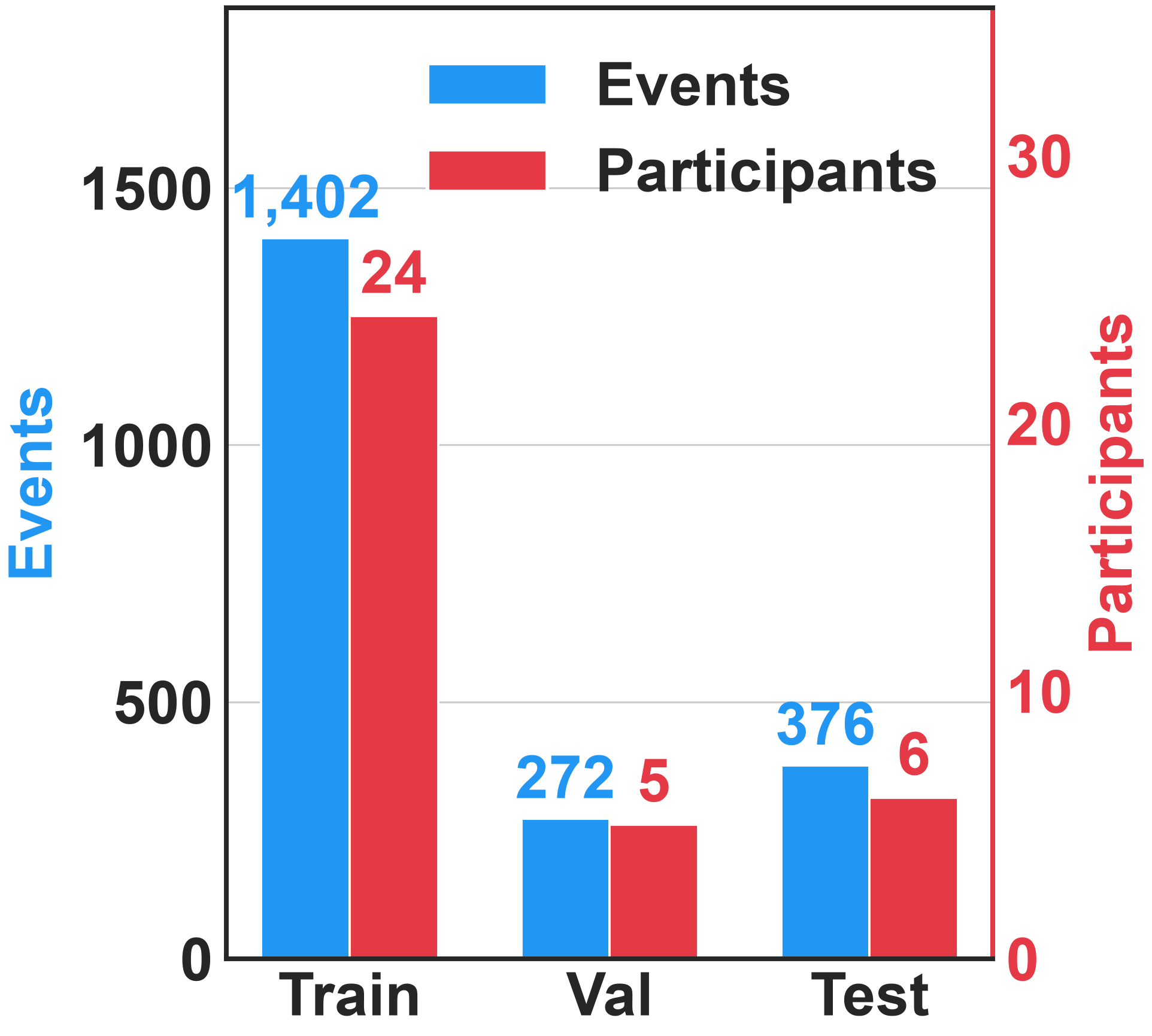}
\caption{Dataset splits}
\label{fig:split_design}
\end{subfigure}
\hfill
\begin{subfigure}[t]{0.305\textwidth}
\centering
\includegraphics[width=\linewidth]{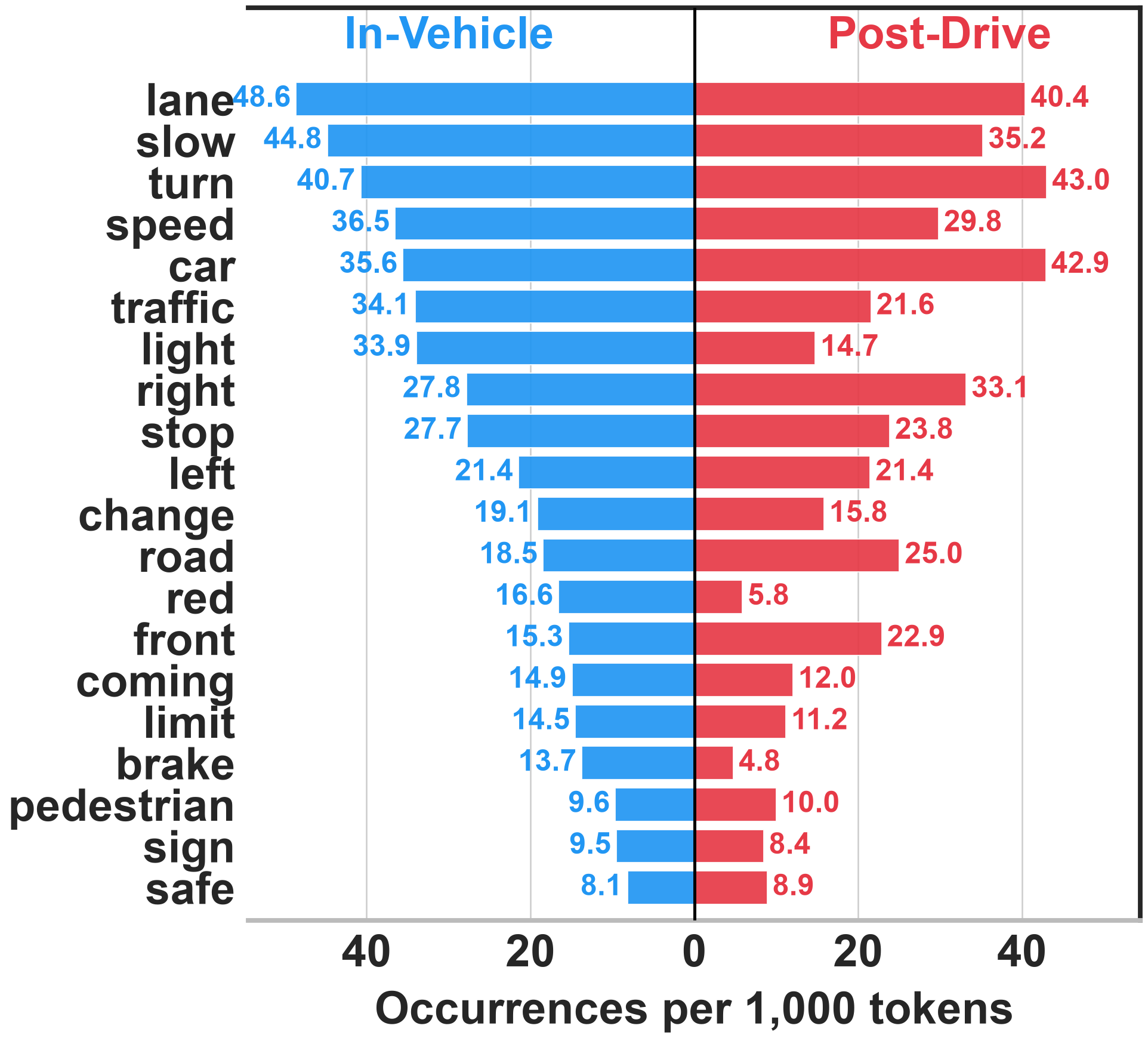}
\caption{Top 20 content words}
\label{fig:top_words}
\end{subfigure}

\caption{Dataset statistics and distributions. Panels summarise the 14.5 km route, action and context labels, train/validation/test splits, reliability-sample representativeness, explanation length, and content-word differences between in-vehicle and post-drive explanations.}
\label{fig:distributions}
\end{figure}

\noindent\textbf{Route, labels, and language statistics.}
Fig.~\ref{fig:distributions} summarises route characteristics, label distributions, split design, reliability-sample representativeness, and explanation-language statistics. The predefined 14.5 km Brisbane route covers major arterial roads, motorway/freeway segments, collector roads, and local streets, with speed zones dominated by 60 km/h and 50 km/h sections. The route also includes common traffic and safety features such as pedestrian crossings, signalised and unsignalised junctions, standalone signs, speed bumps, and stop/give-way signs. Raw transcriptions underwent deterministic rule-based cleaning, including filler removal, capitalisation, punctuation correction, and spell-checking with domain-term protection. Overall, 95.5\% of explanation fields required no correction, and no semantic content was altered. Post-drive explanations are significantly longer than in-vehicle explanations (median 19 vs.\ 13 words. Mann--Whitney $U{=}521{,}171$, $p{<}0.001$), consistent with post-drive reflection allowing more elaboration than real-time narration. Content-word differences in Fig.~\ref{fig:top_words} further show that in-vehicle explanations emphasise immediate spatial and control terms, whereas post-drive explanations more often include referential or scene-reconstruction terms.

\noindent\textbf{Action, context, and SA distributions.}
NARRATE reflects the long-tailed structure of naturalistic driving. Driver actions are imbalanced (Fig.~\ref{fig:action_dist}): \textit{slow down} is the most frequent action (976 events; 47.6\%), followed by \textit{lane change} (451; 22.0\%) and \textit{speed up} (187; 9.1\%). The four non-action classes collectively represent 9.0\% of events. Context labels show a similar long-tailed pattern (Fig.~\ref{fig:fg_context}), with the most frequent fine-grained categories being vehicle following (430; 21.0\%), route-preparation lane change (372; 18.1\%), speed-limit adherence (301; 14.7\%), and traffic-signal compliance (232; 11.3\%). Together, these top four categories account for 65.1\% of events. SA labels are common but not uniform across levels. L2 Comprehension has the highest coverage in both conditions (IV 93.6\%, Post 91.3\%), followed by L1 Perception (IV 89.0\%, Post 83.3\%). L3 Projection is less frequent but appears at similar rates in both explanation timings (IV 69.1\%, Post 70.2\%), indicating that anticipatory reasoning is present even in concise in-vehicle explanations. All three levels co-occur in 59.1\% of in-vehicle and 58.6\% of post-drive explanations.

\noindent\textbf{Explanation trigger source.}
The final event sources were participant think-aloud (804 events), researcher-triggered prompts (385 events), and engineer-flagged deferrals (861 events). Participant think-aloud and researcher-triggered prompts yielded near-complete in-vehicle narration rates (97.4\% and 97.7\%, respectively). In contrast, engineer-flagged deferral events produced in-vehicle explanations for only 13.2\% of rows but post-drive explanations for 95.9\%, reflecting their intended role in capturing events that were unsafe or impractical to discuss during active driving.

  \begin{table}[t]
  \centering
  \footnotesize
  \setlength{\tabcolsep}{10pt}
  \renewcommand{\arraystretch}{0.92}
  \caption{Inter-annotator agreement on the 303-instance reliability subset.
  CI\,=\,bootstrap 95\% confidence interval (1{,}000 resamples);
  Obs.\,=\,observed agreement; Fine/High\,=\,fine-/high-level context labels;
  Exact\,=\,exact-match agreement; Jaccard and $\alpha_\mathrm{MASI}$ measure multi-label overlap.
  SA uses annotators A, B, C; context uses A and C.
  $^\dagger$Prevalence paradox: all annotators ${>}82\%$ positive, so observed agreement is informative.}
  \label{tab:iaa}
  \begin{minipage}[t]{0.64\linewidth}
  \centering
  \textit{SA annotation (A, B, C)}\\[3pt]
  \resizebox{\linewidth}{!}{%
  \begin{tabular}{@{}lccccc@{}}
  \toprule
   & \multicolumn{2}{c}{\textbf{Three-way}} & \multicolumn{2}{c}{\textbf{Pairwise}} & \\
  \cmidrule(lr){2-3}\cmidrule(lr){4-5}
  \textbf{Level} & Fleiss $\kappa$ & 95\% CI & A--C $\kappa$ & A--B $\kappa$ & Obs.\ (A--C) \\
  \midrule
  L1 Perception    & 0.580 & [.49, .67] & 0.820\,[.70,.91] & 0.545\,[.38,.70] & 0.964 \\
  L2 Comp.$^\dagger$ & 0.050 & [-.07,.17] & 0.304\,[.00,.55] & 0.006\,[-.08,.10] & 0.957 \\
  L3 Projection    & 0.203 & [.13, .27] & 0.690\,[.61,.77] & 0.112\,[.08,.15] & 0.858 \\
  \midrule
  Combination      & ---   & ---        & \multicolumn{2}{c}{A--C $\alpha_\mathrm{MASI}{=}0.660\;[.58,.73]$} & --- \\
  \bottomrule
  \end{tabular}}
  \end{minipage}%
  \hfill
  \begin{minipage}[t]{0.33\linewidth}
  \centering
  \textit{Context annotation (A, C)}\\[3pt]
  \begin{tabular}{@{}lcc@{}}
  \toprule
   & \textbf{Fine} & \textbf{High} \\
  \midrule
  Exact \%       & 82.2 & 85.8 \\
  Jaccard        & 0.892 & 0.911 \\
  $\alpha_\mathrm{MASI}$ & 0.852 & 0.861 \\
  95\% CI        & [.81,.89] & [.82,.90] \\
  \bottomrule
  \end{tabular}
  \end{minipage}
  \end{table}

\noindent\textbf{Inter-annotator agreement.}
  Table~\ref{tab:iaa} reports agreement on the 303-instance reliability
  subset (maximum deviation 3.4 pp from the full-dataset distribution;
  Fig.~\ref{fig:sample_dist}). Context labels are robust across both
  granularities (Krippendorff's $\alpha$ with MASI
  distance~\cite{passonneau2006measuring,artstein2008inter}).
  For SA, we report Fleiss' $\kappa$~\cite{fleiss1971measuring} as the
  primary three-way metric; following Landis and
  Koch~\cite{landis1977measurement}, L1 Perception shows
  moderate-to-substantial three-way agreement. L3 Projection is lower,
  driven by Annotator~B's conservative L3 usage (13.2\% positive vs.\
  71.9\% and 59.7\% for A and C), while A--C pairwise agreement remains
  substantial~\cite{artstein2008inter}. L2 is dominated by the prevalence paradox~\cite{feinstein1990high}, so observed three-way agreement (79.2\%) is the relevant metric. About 85\% of released labels are single-annotator (A), with reliability estimated on this subset and supported by substantial A--C agreement at L3.

\section{Benchmarks}
\label{sec:benchmarks}

We define four participant-disjoint baseline tasks to characterise what NARRATE enables and where it remains challenging: Situational Awareness (SA) classification, scenario-context classification, driver-action classification, and explanation generation. These benchmarks support dataset use and comparison rather than proposing a new model, and use the splits in Table~\ref{tab:dataset}.

Unless otherwise stated, text models use a maximum sequence length of 128, batch size 16, AdamW~\cite{loshchilov2017decoupled} with learning rate $2{\times}10^{-5}$ and weight decay 0.01, 5 epochs, linear warmup, and validation macro-F1 for model selection. All trainable baselines were run with five random seeds (1, 2, 3, 4, 42), and results are reported as mean $\pm$ standard deviation. Visual features are extracted using frozen CLIP ViT-B/32~\cite{radford2021learning}: eight frames are sampled uniformly from each 15-second event clip, embedded into 512-dimensional features, and mean-pooled to one event representation. Kinematic features are sampled at the same eight temporal indices from vehicle speed and longitudinal acceleration, yielding a 16-dimensional event vector.

\subsection{T1: Situational Awareness Classification}
\label{sec:t1}

T1 evaluates whether the SA structure annotated in NARRATE is recoverable from driver-produced language. The task is framed as three independent binary classifications for L1 Perception, L2 Comprehension, and L3 Projection, evaluated separately for in-vehicle and post-drive explanations. Models use \texttt{BCEWithLogitsLoss} with threshold 0.5. The primary metric is macro-F1 across the three SA levels. Baselines include majority class, TF-IDF+LR, DistilBERT~\cite{sanh2019distilbert}, BERT-base~\cite{devlin2019bert}, and RoBERTa-base~\cite{liu2019roberta}. Table~\ref{tab:sa_results} shows that the majority baseline is already strong because L1 and L2 have high positive-label prevalence. This indicates that perception and comprehension are frequently expressed in driver explanations, but it also makes these labels less discriminative. L3 Projection provides a more informative test case: it is less frequent and shows clearer model gains over the majority baseline. The most informative result is on L3, where RoBERTa-base beats the majority baseline by the largest margin (0.863 vs.\ 0.773). The high scores on L1 and L2 mostly reflect their high label prevalence rather than learned signal (RoBERTa \ms{0.914}{.007} in-vehicle). For post-drive explanations, DistilBERT gives the highest reported macro-F1 (\ms{0.914}{.007}). Across models, seed variance is low (std $\leq .008$), supporting the view that SA structure is a stable and learnable signal in NARRATE.

\FloatBarrier

\subsection{T2: Driving-Context Classification}
\label{sec:t2}

T2 evaluates whether scenario context can be inferred from explanation text. We consider two granularities: six high-level multi-label categories for in-vehicle and post-drive explanations, and 32 fine-grained multi-label categories using \texttt{primary\_text}. Here, \texttt{primary\_text} denotes the in-vehicle explanation when available, and otherwise the post-drive explanation. For the fine-grained setting, macro-F1 is computed over labels with at least five test instances. The 32-class task is evaluated with BERT-base and RoBERTa-base only. As shown in Table~\ref{tab:context_results}, high-level context is partially recoverable from explanation text. RoBERTa-base performs best at the six-class level for both in-vehicle and post-drive explanations, reaching \ms{0.407}{.023} and \ms{0.418}{.004} macro-F1, respectively. The gap between macro-F1 and weighted-F1 reflects the long-tailed label distribution: common contexts are more learnable, while rare categories remain difficult. The 32-class fine-grained task is substantially harder. RoBERTa-base reaches only \ms{0.035}{.011} macro-F1 and \ms{0.100}{.048} weighted-F1, while BERT-base collapses to near-zero performance on one seed. The near-floor macro-F1 on fine-grained context is itself an informative result: it establishes that 32-class context recognition from explanation text alone is not a solved problem with current text encoders, and that richer grounding in visual scene, map topology, and interaction cues will be necessary, making NARRATE a useful dataset for future multimodal context modelling.

\begin{table}[t]
\caption{T1: Situational Awareness classification on the participant-disjoint test set.
MF1 = Macro-F1; $\uparrow$ higher is better.
Trainable models report mean $\pm$ std over 5 seeds.}
\label{tab:sa_results}
\centering
\footnotesize
\setlength{\tabcolsep}{10pt}
\renewcommand{\arraystretch}{0.92}
\resizebox{\textwidth}{!}{%
\begin{tabular}{@{}lcccccccc@{}}
\toprule
& \multicolumn{4}{c}{\textbf{In-vehicle}}
& \multicolumn{4}{c}{\textbf{Post-drive}} \tabularnewline
\cmidrule(lr){2-5}\cmidrule(lr){6-9}
\textbf{Model} & L1 & L2 & L3 & MF1 & L1 & L2 & L3 & MF1 \tabularnewline
\midrule
Majority
& 0.906 & 0.964 & \worst{0.773} & \worst{0.881}
& \worst{0.904} & 0.946 & \worst{0.801} & \worst{0.884} \tabularnewline
TF-IDF+LR
& 0.908 & 0.964 & 0.811 & 0.894
& 0.916 & 0.949 & 0.857 & 0.907 \tabularnewline
DistilBERT
& 0.906 & 0.964 & 0.846 & \ms{0.905}{.004}
& 0.915 & 0.946 & \best{0.882} & \ms{0.914}{.007} \tabularnewline
BERT-base
& \best{0.917} & \worst{0.963} & 0.854 & \ms{0.911}{.006}
& 0.915 & 0.949 & 0.878 & \ms{0.914}{.005} \tabularnewline
RoBERTa-base
& 0.915 & 0.964 & \best{0.863} & \best{\ms{0.914}{.007}}
& \best{0.924} & 0.949 & 0.861 & \ms{0.911}{.008} \tabularnewline
\bottomrule
\end{tabular}}
\end{table}

\begin{table}[t]
\caption{T2: Context classification on the participant-disjoint test set.
MF1 = Macro-F1; WF1 = Weighted-F1; $\uparrow$ higher is better.
The 32-class MF1 is computed over labels with at least five test instances.
Trainable models report mean $\pm$ std over 5 seeds.}
\label{tab:context_results}
\centering
\footnotesize
\setlength{\tabcolsep}{10pt}
\renewcommand{\arraystretch}{0.92}
\resizebox{\textwidth}{!}{%
\begin{tabular}{@{}lcccccc@{}}
\toprule
& \multicolumn{2}{c}{\textbf{6-class IV}}
& \multicolumn{2}{c}{\textbf{6-class Post}}
& \multicolumn{2}{c}{\textbf{32-class Primary}} \tabularnewline
\cmidrule(lr){2-3}\cmidrule(lr){4-5}\cmidrule(lr){6-7}
\textbf{Model} & MF1 & WF1 & MF1 & WF1 & MF1 & WF1 \tabularnewline
\midrule
Majority
& \worst{0.000} & $-$
& \worst{0.000} & $-$
& $-$ & $-$ \tabularnewline
TF-IDF+LR
& 0.342 & $-$
& 0.377 & $-$
& $-$ & $-$ \tabularnewline
DistilBERT
& \ms{0.315}{.025} & \ms{0.562}{.040}
& \ms{0.309}{.048} & \worst{\ms{0.516}{.076}}
& $-$ & $-$ \tabularnewline
BERT-base
& \ms{0.289}{.045} & \worst{\ms{0.520}{.072}}
& \ms{0.317}{.049} & \ms{0.534}{.079}
& \best{\ms{0.038}{.023}} & \worst{\ms{0.086}{.061}} \tabularnewline
RoBERTa-base
& \best{\ms{0.407}{.023}} & \best{\ms{0.697}{.023}}
& \best{\ms{0.418}{.004}} & \best{\ms{0.703}{.008}}
& \worst{\ms{0.035}{.011}} & \best{\ms{0.100}{.048}} \tabularnewline
\bottomrule
\end{tabular}}
\end{table}

\begin{table}[t]
\caption{T3: Driver-action classification on the participant-disjoint test set ($n{=}376$).
Acc = Accuracy; MF1 = Macro-F1; $\uparrow$ higher is better. Trainable models report mean $\pm$ std over 5 seeds.}
\label{tab:action_results}
\centering
\footnotesize
\setlength{\tabcolsep}{10pt}
\renewcommand{\arraystretch}{0.92}
\begin{tabular}{@{}llcc@{}}
\toprule
\textbf{Model} & \textbf{Modality} & \textbf{Acc} & \textbf{MF1} \tabularnewline
\midrule
Majority        & $-$          & 0.476                    & \worst{0.065} \tabularnewline
TF-IDF+LR       & Text         & 0.681                    & 0.228 \tabularnewline
CLIP+LR         & Video        & \worst{0.468}            & 0.219 \tabularnewline
Kinematics MLP  & Motion       & \ms{0.586}{.007}         & \ms{0.241}{.007} \tabularnewline
CLIP+Kin MLP    & Video+Motion & \ms{0.629}{.015}         & \best{\ms{0.306}{.016}} \tabularnewline
DistilBERT      & Text         & \ms{0.708}{.018}         & \ms{0.246}{.024} \tabularnewline
BERT-base       & Text         & \ms{0.690}{.034}         & \ms{0.228}{.028} \tabularnewline
RoBERTa-base    & Text         & \best{\ms{0.728}{.009}}  & \ms{0.289}{.018} \tabularnewline
BERT+Kin        & Text+Motion  & \ms{0.725}{.011}         & \ms{0.280}{.007} \tabularnewline
\bottomrule
\end{tabular}
\end{table}

\begin{table}[t]
\caption{T4: Explanation generation on the participant-disjoint test set ($n{=}376$).
B4 = BLEU-4; RL = ROUGE-L; MET = METEOR; BS = BERTScore-F1;
$\uparrow$ higher is better. Fine-tuned models report mean $\pm$ std over 5 seeds.}
\label{tab:generation_results}
\centering
\footnotesize
\setlength{\tabcolsep}{10pt}
\renewcommand{\arraystretch}{0.92}
\resizebox{\textwidth}{!}{%
\begin{tabular}{@{}llcccc@{}}
\toprule
\textbf{Model} & \textbf{Type} & \textbf{B4} & \textbf{RL} & \textbf{MET} & \textbf{BS} \tabularnewline
\midrule
Random
& Retrieval
& \ms{0.010}{.001} & \ms{0.115}{.005} & \ms{0.091}{.004} & \ms{0.872}{.001} \tabularnewline
Action
& Retrieval
& \ms{0.012}{.004} & \ms{0.129}{.028} & \ms{0.115}{.047} & \ms{0.872}{.003} \tabularnewline
Action+Context
& Retrieval
& \ms{0.019}{.007} & \ms{0.158}{.016} & \ms{0.133}{.021} & \ms{0.879}{.005} \tabularnewline
CLIP-NN
& Retrieval
& 0.012 & 0.132 & 0.108 & 0.875 \tabularnewline
CLIP+Kin-NN
& Retrieval
& 0.013 & 0.142 & 0.110 & 0.876 \tabularnewline
FLAN-T5-base
& Fine-tuned
& \ms{0.034}{.001} & \ms{0.215}{.010} & \ms{0.144}{.019} & \ms{0.892}{.001} \tabularnewline
T5-base
& Fine-tuned
& \best{\ms{0.036}{.002}} & \best{\ms{0.238}{.010}} & \ms{0.176}{.010} & \best{\ms{0.895}{.003}} \tabularnewline
BART-base
& Fine-tuned
& \ms{0.024}{.009} & \ms{0.169}{.033} & \ms{0.112}{.038} & \ms{0.894}{.003} \tabularnewline
GPT-2
& Fine-tuned
& \ms{0.030}{.002} & \ms{0.236}{.013} & \best{\ms{0.185}{.010}} & \ms{0.892}{.004} \tabularnewline
Phi-3-mini
& 0-shot
& \worst{0.004} & \worst{0.063} & \worst{0.083} & \worst{0.815} \tabularnewline
\bottomrule
\end{tabular}}
\end{table}

\subsection{T3: Driver Action Classification}
\label{sec:t3}

T3 evaluates whether driver actions can be inferred from text, video, motion, or their combinations. The task is a 10-class single-label classification problem, evaluated primarily with macro-F1 and secondarily with accuracy. We compare text baselines, frozen CLIP visual features, kinematic features, and simple fusion baselines: \textit{Kinematics MLP} uses speed and acceleration, while \textit{CLIP+Kin MLP} and \textit{BERT+Kin} concatenate kinematic features with frozen CLIP video features and BERT text embeddings, respectively. Table~\ref{tab:action_results} shows that text is the strongest single modality: RoBERTa-base achieves the highest text-only accuracy (\ms{0.728}{.009}) and macro-F1 (\ms{0.289}{.018}). Kinematic features also provide useful signal, outperforming frozen CLIP video features in macro-F1. The best macro-F1 overall is obtained by the CLIP+Kin MLP (\ms{0.306}{.016}), suggesting that visual and motion cues are complementary for action recognition even when visual features are frozen and simple. In contrast, BERT+Kin achieves strong accuracy and low variance but does not improve macro-F1 over RoBERTa text alone, indicating that simple fusion is not sufficient to resolve rare classes. Overall performance remains limited by class imbalance: dominant actions such as \textit{slow down} and \textit{lane change} are easier to classify, while rare and non-action classes drive the macro-F1 shortfall.

\FloatBarrier

\subsection{T4: Explanation Generation}
\label{sec:t4}

T4 evaluates the difficulty of generating natural-language driver explanations from structured event information. This task is intended as a structured-input generation baseline, not as an end-to-end multimodal generation model. Each fine-tuned generator conditions on the action label, context category, and kinematic trace. Retrieval baselines provide additional comparisons using random selection, label matching, frozen visual nearest neighbours, and visual+kinematic nearest neighbours. Fine-tuned generators include FLAN-T5-base~\cite{chung2024scaling}, T5-base~\cite{raffel2020exploring}, BART-base~\cite{lewis2020bart}, and GPT-2~\cite{radford2019language}; Phi-3-mini-4k-instruct~\cite{abdin2024phi3} is evaluated zero-shot. We evaluate BLEU-4~\cite{papineni2002bleu}, ROUGE-L~\cite{lin2004rouge}, METEOR~\cite{banerjee2005meteor}, and BERTScore-F1~\cite{zhang2019bertscore}. Table~\ref{tab:generation_results} shows that fine-tuned generators outperform retrieval baselines on most metrics. Among retrieval methods, action+context matching performs best, indicating that symbolic event labels capture more explanation similarity than frozen visual nearest neighbours alone. T5-base achieves the best scores on three metrics: BLEU-4
(\ms{0.036}{.002}), ROUGE-L (\ms{0.238}{.010}), and
BERTScore-F1 (\ms{0.895}{.003}). GPT-2 achieves the best
METEOR score, with \ms{0.185}{.010}. BART-base shows higher variance than the other fine-tuned models, indicating sensitivity to random initialisation. Zero-shot Phi-3-mini underperforms all fine-tuned models, suggesting a domain gap between generic instruction following and naturalistic driver explanations. The modest absolute scores are important for interpreting the benchmark. NARRATE explanations are concise, diverse, and context-specific. Multiple valid explanations may describe the same event using different wording or levels of detail. As a result, exact n-gram overlap remains low even when generated explanations are semantically plausible. T4 establishes a text-conditioned lower bound and confirms that naturalistic driver explanation generation remains unsolved under current paradigms, which is an open problem the dataset is designed to support, not one it is expected to solve.

\FloatBarrier


\section{Limitations}
\label{sec:limitations}

NARRATE is modest in scale, limited to a single daytime route in Brisbane, and subject to long-tailed action and context distributions that leave rare classes thin. The baselines use simplified event-level representations as a deliberate lower bound. Full exploitation of the temporal, LiDAR, and multi-view streams is left to future work.

\section{Conclusion}
We introduced NARRATE, a multimodal real-world Australian driving dataset for human-centred explanations in automated driving. NARRATE contains 2,050 annotated events from 35 experienced drivers and driving instructors on public roads, pairing synchronised visual, LiDAR, localisation, inertial, and kinematic streams with in-vehicle and post-drive explanations produced by the drivers themselves. It provides driver-action labels, scenario-context labels, and span-level Situational Awareness annotations for Perception, Comprehension, and Projection. Baselines show that SA structure is learnable from driver language, while fine-grained context recognition and naturalistic explanation generation remain challenging. NARRATE offers a domain-aware dataset and evaluation testbed for models that reflect how human drivers perceive, interpret, and anticipate driving situations.

\section*{Data Availability}
NARRATE is available through mediated access via the QUT Research Data Finder at
\url{https://doi.org/10.25912/RDF_1786669427992}. Dataset documentation and updates are available at \url{https://github.com/ashkan-zadeh/NARRATE}.

\section*{Acknowledgements}
This research was supported by the Australian Research Council Discovery Project (DP220102598).

\bibliographystyle{unsrt}
\bibliography{references}

\end{document}